\documentclass[english]{article}
\usepackage[T1]{fontenc}
\usepackage[latin9]{inputenc}
\usepackage{amsbsy}
\usepackage{amstext}
\usepackage{amssymb}
\usepackage{graphicx}

\makeatletter

\providecommand{\tabularnewline}{\\}

\@ifundefined{definecolor}
 {\usepackage{color}}{}
\usepackage{aaai17}\usepackage{times}\usepackage{helvet}\usepackage{courier}

\usepackage{flushend}

\usepackage[caption=false,font=footnotesize]{subfig}
\usepackage{microtype}
\usepackage{flushend}

\usepackage{amsbsy}
\usepackage{algorithmic}
\usepackage[english]{babel}
\usepackage[T1]{fontenc}
\makeatother

\usepackage{babel}
\begin{document}
\global\long\def\model{\mathsf{\mathtt{CLN}}}

\title{Column Networks for Collective Classification}

\author{Trang Pham, Truyen Tran, Dinh Phung and Svetha Venkatesh\\
 Deakin University, Australia\\
\{\emph{phtra,truyen.tran,dinh.phung,svetha.venkatesh}\}\emph{@deakin.edu.au}}
\maketitle
\begin{abstract}
Relational learning deals with data that are characterized by relational
structures. An important task is collective classification, which
is to jointly classify networked objects. While it holds a great promise
to produce a better accuracy than non-collective classifiers, collective
classification is computationally challenging and has not leveraged
on the recent breakthroughs of deep learning. We present Column Network
($\model$), a novel deep learning model for collective classification
in multi-relational domains. $\model$ has many desirable theoretical
properties: (i) it encodes multi-relations between any two instances;
(ii) it is deep and compact, allowing complex functions to be approximated
\emph{at the network level} with a small set of free parameters; (iii)
local and relational features are learned simultaneously; (iv) long-range,
higher-order dependencies between instances are supported naturally;
and (v) crucially, learning and inference are efficient with linear
complexity in the size of the network and the number of relations.
We evaluate $\model$ on multiple real-world applications: (a) delay
prediction in software projects, (b) PubMed Diabetes publication classification
and (c) film genre classification. In all of these applications, $\model$
demonstrates a higher accuracy than state-of-the-art rivals.
\end{abstract}
\global\long\def\xb{\boldsymbol{x}}
\global\long\def\yb{\boldsymbol{y}}
\global\long\def\eb{\boldsymbol{e}}
\global\long\def\zb{\boldsymbol{z}}
\global\long\def\hb{\boldsymbol{h}}
\global\long\def\ab{\boldsymbol{a}}
\global\long\def\bb{\boldsymbol{b}}
\global\long\def\cb{\boldsymbol{c}}
\global\long\def\sigmab{\boldsymbol{\sigma}}
\global\long\def\gammab{\boldsymbol{\gamma}}
\global\long\def\alphab{\boldsymbol{\alpha}}
\global\long\def\rb{\boldsymbol{r}}
\global\long\def\fb{\boldsymbol{f}}
\global\long\def\ib{\boldsymbol{i}}
\global\long\def\thetab{\boldsymbol{\theta}}
\global\long\def\pb{\boldsymbol{p}}

\section{Introduction}

Relational data are characterized by relational structures between
objects or data instances. For example, research publications are
linked by citations, web pages are connected by hyperlinks and movies
are related through same directors or same actors. Using relations
may improve performance in classification as relations between entities
may be indicative of relations between classes. A canonical task in
learning from this data type is \emph{collective classification} in
which networked data instances are classified simultaneously rather
than independently to exploit the dependencies in the data \cite{macskassy2007cnd,neville2007relational,Richardson-Domingos-ML06,sen2008collective}.
Collective classification is, however, highly challenging. Exact collective
inference under general dependencies is intractable. For tractable
learning, we often resort to surrogate loss functions such as (structured)
pseudo-likelihood \cite{Sutton-McCallum-ICML07}, approximate gradient
\cite{Hinton02}, or iterative schemes, stacked learning \cite{choetkiertikul2015predicting,kou2007stacked,macskassy2007cnd,neville2000iterative}.

Existing models designed for collective classification are mostly
shallow and do not emphasize learning of local and relational features.
Deep neural networks,\emph{ }on the other hand, offer automatic feature
learning, which is arguably the key behind recent record-breaking
successes in vision, speech, games and NLP \cite{lecun2015deep,mnih2015human}.
With known challenges in relational learning, \emph{can we design
a deep neural network that is efficient and accurate for collective
classification?} There has been recent work that combines deep learning
with structured prediction but the main learning and inference problems
for general multi-relational settings remain open \cite{belanger2015structured,do2010neural,tompson2014joint,yu2010sequential,zheng2015conditional}.

In this paper, we present \emph{Column Network} ($\model$), an efficient
deep learning model for multi-relational data, with emphasis on collective
classification. The design of $\model$ is partly inspired by the
columnar organization of neocortex \cite{mountcastle1997columnar},
in which cortical neurons are organized in vertical, layered mini-columns,
each of which is responsible for a small receptive field. Communications
between mini-columns are enabled through short-range horizontal connections.
In $\model$, each mini-column is a feedforward net that takes an
input vector \textendash{} which plays the role of a receptive field
\textendash{} and produces an output class. Each mini-column net not
only learns from its own data but also exchanges features with neighbor
mini-columns along the pathway from the input to output. Despite the
short-range exchanges, the interaction range between mini-columns
increases with depth, thus enabling long-range dependencies between
data objects.

To be able to learn with hundreds of layers, we leverage the recently
introduced highway nets \cite{srivastava2015training} as models for
mini-columns. With this design choice, $\model$ becomes a \emph{network
of interacting highway nets}. But unlike the original highway nets,
$\model$'s hidden layers share the same set of parameters, allowing
the depth to grow without introducing new parameters \cite{liao2016bridging,pham2016faster}.
Functionally, if feedforward nets and highway nets are functional
approximators for an input vector, $\model$ can be thought as an
approximator of a grand function that takes a complex network of vectors
as input and returns multiple outputs. $\model$ has many desirable
theoretical properties: (i) it encodes multi-relations between any
two instances; (ii) it is deep and compact, allowing complex functions
to be approximated \emph{at the network level} with a small set of
free parameters; (iii) local and relational features are learned simultaneously;
(iv) long-range, higher-order dependencies between instances are supported
naturally; and (v) crucially, learning and inference are efficient,
linear in the size of network and the number of relations.

We evaluate $\model$ on real-world applications: (a) delay prediction
in software projects, (b) PubMed Diabetes publication classification
and (c) film genre classification. In all applications, $\model$
demonstrates a higher accuracy than state-of-the-art rivals.

\section{Preliminaries}

\emph{Notation convention:} We use capital letters for matrices and
bold lowercase letters for vectors. The sigmoid function of a scalar
$x$ is defined as $\sigma(x)=\left[1+\text{exp}(-x)\right]^{-1}$,
$x\in\mathbb{R}$. A function $g$ of a vector $\xb$ is defined as
$g(\xb)=\left(g(x_{1}),...,g(x_{n})\right)$. The operator $*$ is
used to denote element-wise multiplication. We use superscript $t$
(e.g. $\hb^{t}$) to denote layers or computational steps in neural
networks , and subscript $i$ for the $i^{th}$ element in a set (e.g.
$\hb_{i}^{t}$ is the hidden activation at layer $t$ of entity $i$
in a graph). 

\subsection{Collective Classification in Multi-relational Setting \label{subsec:Collective-Classification}}

We describe the collective classification setting under multiple relations.
Given a graph of entities $\text{\textbf{G}=\{\textbf{E}, \textbf{R}, \textbf{X}, \textbf{Y}\}}$
where $\text{\textbf{E}}=\{\eb_{1},...,\eb_{N}\}$ are $N$ entities
that connect through relations in $\text{\textbf{R}}$. Each tuple
$\{\eb_{j},\eb_{i},r\}\in\text{\textbf{R}}$ describes a relation
of type $r$ ($r=1...R,$ where $R$ is the number of relation types
in $\text{\textbf{G}}$) from entity $\eb_{j}$ to entity $\eb_{i}$.
Two entities can connect through multiple relations. A relation can
be unidirectional or bidirectional. For example, movie \texttt{A}
and movie \texttt{B} may be linked by a unidirectional relation \texttt{sequel(A,B)}
and two bidirectional relations: \texttt{same-actor(A,B)} and \texttt{same-director(A,B)}.

Entities and relations can be represented in an entity graph where
a node represents an entity and an edge exists between two nodes if
they have at least one relation. Furthermore, $\eb_{j}$ is a neighbor
of $\eb_{i}$ if there is a link from $\eb_{j}$ to $\eb_{i}$. Let
$\mathcal{N}(i)$ be the set of all neighbors of $\eb_{i}$ and $\mathcal{N}_{r}(i)$
be the set of neighbors related to $\eb_{i}$ through relation $r$.
This immediately implies $\mathcal{N}(i)=\cup_{r\in R}\mathcal{N}_{r}(i)$. 

$\text{\textbf{X}}=\{\xb_{1},...,\xb_{N}\}$ is the set of local features,
where $\xb_{i}$ is feature vector of entity $\eb_{i}$; and $\text{\textbf{Y}}=\{y_{1},...,y_{N}\}$
with each $y_{i}\in\{1,...,L\}$ is the label of $\eb_{i}$. $y_{i}$
can either be observed or latent. Given a set of known label entities
$\text{\textbf{E}}_{obs}$, a collective classification algorithm
simultaneously infers unknown labels of entities in the set $\text{\textbf{E}}_{hid}=\text{\textbf{E}}\backslash\text{\textbf{E}}_{obs}$.
In our probabilistic setting, we assume the classifier produces estimate
of the joint conditional distribution $P\left(\textbf{Y}\mid\text{\textbf{G}}\right)$.

It is challenging to learn and infer about $P\left(\textbf{Y}\mid\text{\textbf{G}}\right)$.
A popular strategy is to employ approximate but efficient iterative
methods \cite{macskassy2007cnd}. In the next subsection, we describe
a highly effective strategy known as \emph{stacked learning}, which
partly inspires our work.

\subsection{Stacked Learning \label{subsec:Stacked-Learning}}

\begin{figure}[h]
\centering{}\includegraphics[bb=230bp 153bp 750bp 415bp,clip,width=0.9\columnwidth]{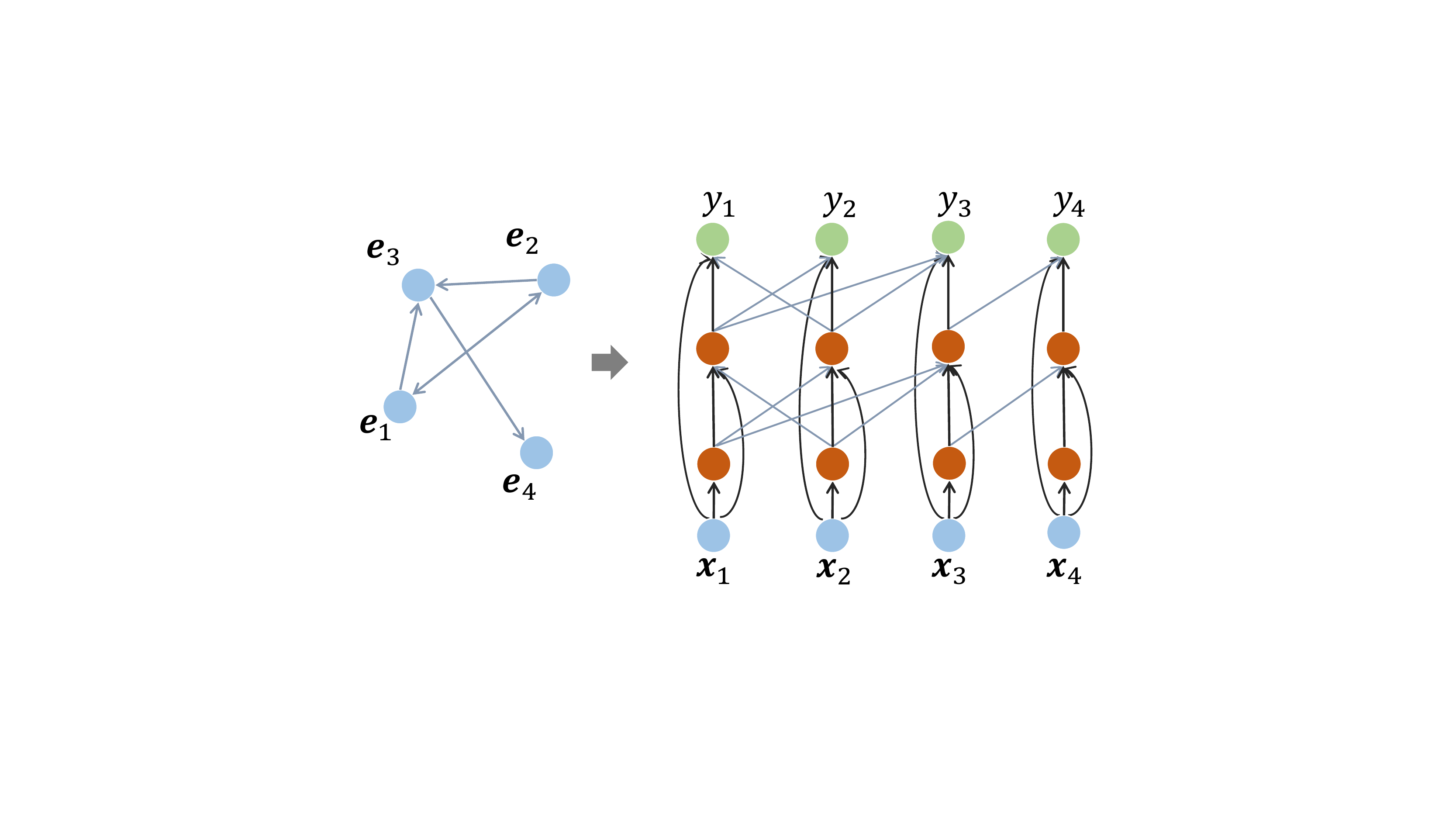}\caption{Collective classification with Stacked Learning (SL). (\textbf{Left}):
A graph with 4 entities connected by unidirectional and bidirectional
links, (\textbf{Right}): SL model for the graph with three steps where
$\protect\xb_{i}$ is the feature vector of entity $\protect\eb_{i}$.
The bidirectional link between $\protect\eb_{1}$ and $\protect\eb_{2}$
is modeled as two unidirectional links from $\protect\eb_{1}$ to
$\protect\eb_{2}$ and vice versa. \label{fig:SI}}
\end{figure}

Stacked learning (Fig.~\ref{fig:SI}) is a multi-step learning procedure
for collective classification \cite{choetkiertikul2015predicting,kou2007stacked,yu2010sequential}.
At step $t-1$, a classifier is used to predict class probabilities
for entity $\eb_{j}$, i.e., $\pb_{j}^{t-1}=\left[P^{t-1}\left(y_{j}=1\right),...,P^{t-1}\left(y_{j}=L\right)\right]$.
These intermediate outputs are then used as relational features for
neighbor classifiers in the next step. In \cite{choetkiertikul2015predicting},
each relation produces one set of \emph{contextual features}, where
all features of the same relation are averaged:

\begin{equation}
\cb_{ir}^{t}=\frac{1}{\left|\mathcal{N}_{r}(i)\right|}\sum_{j\in\mathcal{N}_{r}(i)}\pb_{j}^{t-1}\label{eq:SI-context}
\end{equation}
where $\cb_{ir}^{t}$ is the relational feature vector for relation
$r$ at step $t$. The output at step $t$ is predicted as follows
\begin{equation}
P^{t}\left(y_{i}\right)=f^{t}\left(\xb_{i},\pb_{i}^{t-1},\left[\cb_{i1}^{t},\cb_{i2}^{t}...,\cb_{iR}^{t}\right]\right)\label{eq:SI}
\end{equation}
where $f^{t}$ is the classifier at step $t$. When $t=1$, the model
uses local features of entities for classification, i.e., $\cb_{ir}^{1}=\boldsymbol{0}$
and $\pb_{i}^{0}=\boldsymbol{0}$. At each step, classifiers are trained
sequentially with known-label entities.

\section{Column Networks \label{sec::R2NN}}

In this section we present our main contribution, the Column Network
($\model$).

\subsection{Architecture \label{subsec:CLN-architecture}}

\begin{figure}[h]
\begin{centering}
\includegraphics[bb=390bp 165bp 780bp 410bp,clip,width=0.7\columnwidth]{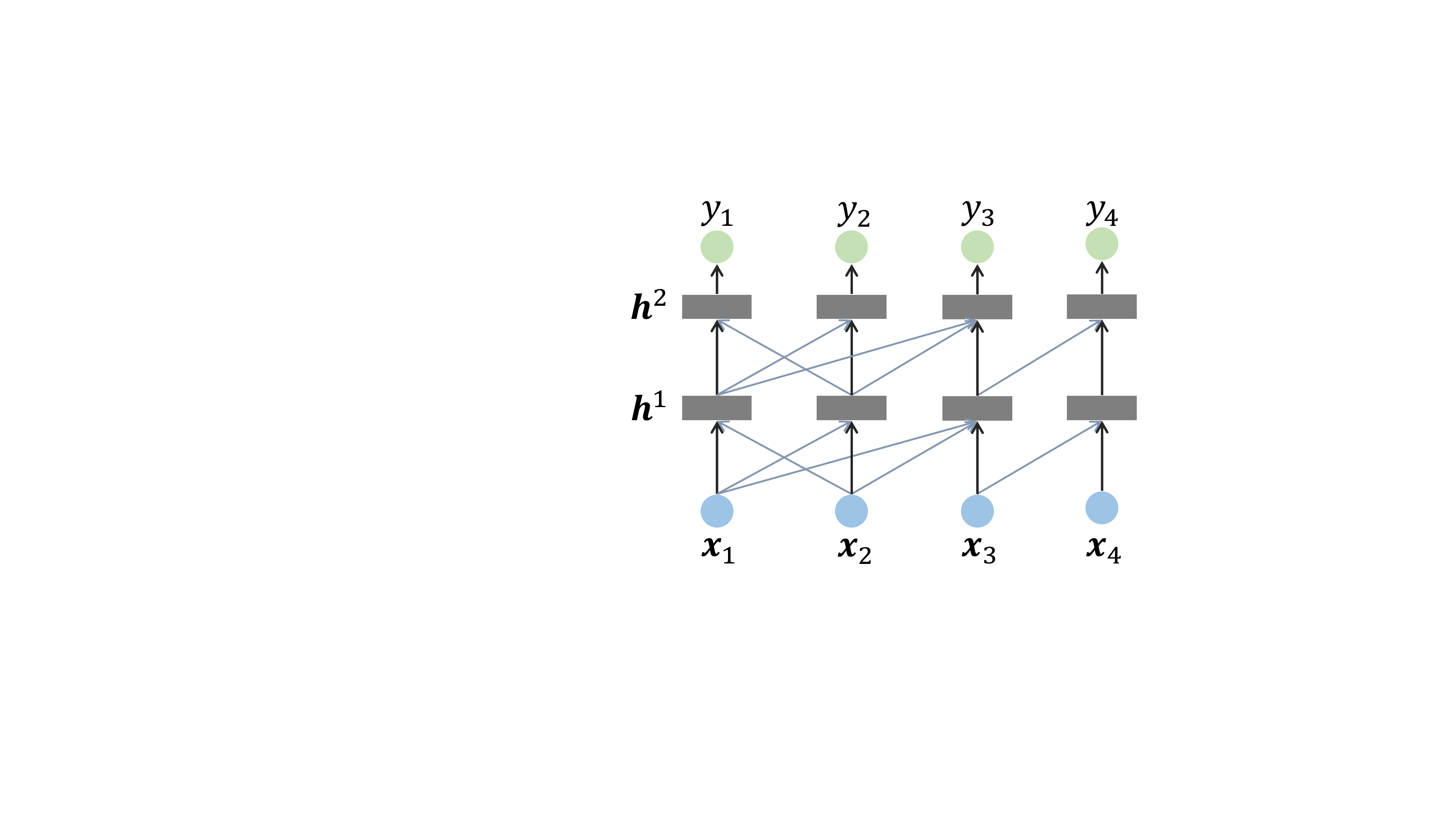}
\par\end{centering}
\caption{$\protect\model$ for the graph in Fig.~\ref{fig:SI}(\textbf{Left})
with 2 hidden layers ($\protect\hb^{1}$ and $\protect\hb^{2}$).
\label{fig:CLN-model}}
\end{figure}

Inspired by the columnar organization in neocortex \cite{mountcastle1997columnar},
the $\model$ has one mini-column per entity (or data instance), which
is akin to a sensory receptive field. Each column is a feedforward
net that passes information from a lower layer to a higher layer of
its own, and higher layers of neighbors (see Fig.~\ref{fig:CLN-model}
for a $\model$ that models the graph in Fig.\ref{fig:SI}(\textbf{Left})).
The nature of the inter-column communication is dictated by the relations
between the two entities. 

Through multiple layers, long-range dependencies are established (see
Sec.~\ref{subsec:Effectiveness-of-CLN} for more in-depth discussion).
This somewhat resembles the strategy used in stacked learning as described
in Sec.~\ref{subsec:Stacked-Learning}. \emph{The main difference
is that in $\model$ the intermediate steps do not output class probabilities
but learn higher abstraction of instance features and relational features}.
As such, our model is \emph{end-to-end} in the sense that receptive
signals are passed from the bottom to the top, and abstract features
are inferred along the way. Likewise, the training signals are passed
from the top to the bottom. 

Denote by $\xb_{i}\in\mathbb{R}^{M}$ and $\hb_{i}^{t}\in\mathbb{R}^{K_{t}}$
the input feature vector and the hidden activation at layer $t$ of
entity $\eb_{i}$, respectively. If there is a connection from entity
$\eb_{j}$ to $\eb_{i}$, $\hb_{j}^{t-1}$ serves as an input for
$\hb_{i}^{t}$ . Generally, $\hb_{i}^{t}$ is a non-linear function
of $\hb_{i}^{t-1}$ and previous hidden states of its neighbors:

\[
\hb_{i}^{t}=g\left(\hb_{i}^{t-1},\hb_{j_{1}}^{t-1},...,\hb_{j_{|N(i)|}}^{t-1}\right)
\]
 where $j\in\mathcal{N}(i)$ and $\hb_{i}^{0}$ is the input vector
$\xb_{i}$.

We borrow the idea of stacked learning (Sec.~\ref{subsec:Stacked-Learning})
to handle multiple relations in $\model$. The context of relation
$r$ ($r=1,...,R)$ at layer $t$ in Eq.~(\ref{eq:SI-context}) is
replaced by
\begin{equation}
\cb_{ir}^{t}=\frac{1}{\left|\mathcal{N}_{r}(i)\right|}\sum_{j\in\mathcal{N}_{r}(i)}\hb_{j}^{t-1}\label{eq:CLN-context}
\end{equation}
Furthermore, different from stacked learning, the context in $\model$
are abstracted features, i.e., we replace Eq.~(\ref{eq:SI}) by
\begin{equation}
\hb_{i}^{t}=g\left(\bb^{t}+W^{t}\hb_{i}^{t-1}+\frac{1}{z}\sum_{r=1}^{R}V_{r}^{t}\cb_{jr}^{t}\right)\label{eq:CLN-hidden}
\end{equation}
where $W^{t}\in\mathbb{R}^{K^{t}\times K^{t-1}}$ and $V_{r}^{t}\in\mathbb{R}^{K^{t}\times K^{t-1}}$
are weight matrices and $\bb^{t}$ is a bias vector for some activation
function $g$; $z$ is a pre-defined constant which is used to prevent
the sum of parameterized contexts from growing too large for complex
relations. 

At the top layer $T$, for example, the label probability for entity
$i$ is given as:
\[
P\left(y_{i}=l\right)=\mbox{softmax}\left(\boldsymbol{b}_{l}+W_{l}\hb_{i}^{T}\right)
\]

\paragraph{Remark:}

There are several similarities between $\model$ and existing neural
network operations. Eq.~(\ref{eq:CLN-context}) implements mean-pooling,
the operation often seen in CNN. The main difference with the standard
CNN is that the mean pooling does not reduce the graph size. This
suggests other forms of pooling such as max-pooling or sum-pooling.
Asymmetric pooling can also be implemented based on the concept of
attention, that is, Eq.~(\ref{eq:CLN-context}) can be replaced by:
\[
\cb_{ir}^{t}=\sum_{j\in\mathcal{N}_{r}(i)}\alpha_{j}\hb_{j}^{t-1}
\]
subject to $\sum_{j\in\mathcal{N}_{r}(i)}\alpha_{j}=1$ and $\alpha_{j}\ge0$.

Eq.~(\ref{eq:CLN-hidden}) implements a convolution. For example,
standard 3x3 convolutional kernels in images implement 8 relations:
\emph{left, right, above, below, above-left, above-right, below-left,
below-right}. Supposed that the relations are shared between nodes,
the $\model$ achieves \emph{translation invariance}, similar to that
in CNN.

\subsection{Highway Network as Mini-Column \label{subsec:Highway-Network-as-Column}}

We now specify the detail of a mini-column, which we implement by
extending a recently introduced feedforward net called Highway Network
\cite{srivastava2015training}. Recall that traditional feedforward
nets have a major difficulty of learning with high number of layers.
This is due to the nested non-linear structure that prevents the ease
of passing information and gradient along the computational path.
Highway nets solve this problem by partially opening the gate that
lets previous states to propagate through layers, as follows:

\begin{equation}
\hb^{t}=\alphab_{1}*\tilde{\hb}^{t}+\alphab_{2}*\hb^{t-1}\label{eq:h_highway}
\end{equation}
where $\tilde{\hb}^{t}$ is a nonlinear candidate function of $\hb^{t-1}$
and where $\alphab_{1},\alphab_{2}\in(\boldsymbol{0},\boldsymbol{1})$
are learnable gates. Since the gates are never shut down completely,
data signals and error gradients can propagate very far in a deep
net.

For modeling relations, the candidate function $\tilde{\hb}^{t}$
in Eq.~(\ref{eq:h_highway}) is computed using Eq.~(\ref{eq:CLN-hidden}).
Likewise, the gates are modeled as:

\begin{eqnarray}
\alphab_{1} & = & \sigma\left(\bb_{\alphab}^{t}+W_{\text{\ensuremath{\alphab}}}^{t}\hb_{i}^{t-1}+\frac{1}{z}\sum_{r=1}^{R}V_{\alphab r}^{t}\cb_{jr}^{t}\right)\label{eq:CLN-gate}
\end{eqnarray}
and $\alphab_{2}=\boldsymbol{1}-\alphab_{1}$ as for compactness \cite{srivastava2015training}.
Other gating options exists, for example, the $p$-norm gates where
$\alphab_{1}^{p}+\alphab_{2}^{p}=\boldsymbol{1}$ for $p>0$ \cite{pham2016faster}. 

\subsection{Parameter Sharing for Compactness \label{subsec:Parameter-Sharing-for-Compactness}}

For feedforward nets, the number of parameters grow with number of
hidden layers. In $\model$, the number is multiplied by the number
of relations (see Eq.~(\ref{eq:CLN-hidden})). In highway network
implementation of mini-columns, a set of parameters for the gates
is used thus doubling the number of parameters (see Eq.~(\ref{eq:CLN-gate})).
For a deep $\model$ with many relations, the number of parameters
may grow faster than the size of training data, leading to overfitting
and a high demand of memory. To address this challenge, we borrow
the idea of parameter sharing in Recurrent Neural Network (RNN), that
is, layers have identical parameters. There has been empirical evidence
supporting this strategy in non-relational data \cite{liao2016bridging,pham2016faster}.

With parameter sharing, the depth of the $\model$ can grow without
increasing in model size. This may lead to good performance on small
and medium datasets. See Sec.~\ref{sec:Experiments-and-Results}
provides empirical evidences.

\subsection{Capturing Long-range Dependencies \label{subsec:Effectiveness-of-CLN}}

An important property of our proposed deep $\model$ is the ability
to capture long-range dependencies despite only local state exchange
as shown in Eqs\@.~(\ref{eq:CLN-hidden},\ref{eq:CLN-gate}). To
see how, let us consider the example in Fig.~\ref{fig:CLN-model},
where $\xb_{1}$ is modeled in $\hb_{3}^{1}$ and $\hb_{3}^{1}$ is
modeled in $\hb_{4}^{2}$, therefore although $\eb_{1}$ does not
directly connect to $\eb_{4}$ but information of $\eb_{1}$ is still
embedded in $\hb_{4}^{2}$ through $\hb_{3}^{1}$. More generally,
after $k$ hidden layers, a hidden activation of an entity can contains
information of its expanded neighbors of radius $k$. When the number
of layers is large, the representation of an entity at the top layer
contains not only its local features and its directed neighbors, but
also the information of the entire graph. With highway networks, all
of these levels of representations are accumulated through layers
and used to predict output labels. 

\subsection{Training with mini-batch \label{subsec:Training-with-mini-batch}}

As described in Sec.~\ref{subsec:CLN-architecture}, $\hb_{i}^{t}$
is a function of $\hb_{i}^{t-1}$ and the previous layer of its neighbors.
$\hb_{i}^{t}$ therefore can contains information of the entire graph
if the network is deep enough. This requires full-batch training which
is expensive and not scalable. We propose a very simple yet efficient
approximation method that allows mini-batch training. For each mini-batch,
the neighbor activations are temporarily frozen to scalars, i.e.,
gradients are not propagated through this ``blanket''. After the
parameter update, the activations are recomputed as usual. Experiments
showed that the procedure did converge and its performance is comparative
with the full-batch training method.

\section{Experiments and Results \label{sec:Experiments-and-Results}}

In this section, we report three real-world applications of $\model$
on networked data: \emph{software delay estimate}, \emph{PubMed paper
classification} and \emph{film genre classification}.

\subsection{Baselines}

For comparison, we employed a comprehensive suit of baseline methods
which include: (a) those designed for collective classification, and
(b) deep neural nets for non-collective classification. For the former,
we used NetKit\footnote{http://netkit-srl.sourceforge.net/}, an open
source toolkit for classification in networked data \cite{macskassy2007cnd}.
NetKit offers a classification framework consisting of 3 components:
a \emph{local classifier}, a \emph{relational classifier} and a \emph{collective
inference} method. In our experiments, the local classifier is the
Logistic Regression (LR) for all settings; relational classifiers
are (i) weighted-vote Relational Neighbor (wvRN), (ii) logistic regression
link-based classifier with normalized values (nbD), and (iii) logistic
regression link-based classifier with absolute count values (nbC).
Collective inference methods include Relaxation Labeling (RL) and
Iterative Classification (IC). In total, there are 6 pairs of ``relational
classifier \textendash{} collective inference'': wvRN-RL, wvRN-IC,
nbD-RL, nbD-IC, nbC-RL and nbC-IC. For each dataset, results of two
best settings will be reported.

We also implemented the state-of-the-art collective classifiers following
\cite{choetkiertikul2015predicting,kou2007stacked,yu2010sequential}:
stacked learning with logistic regression (SL-LR) and with random
forests (SL-RF).

For deep neural nets, following the latest results in \cite{liao2016bridging,pham2016faster},
we implemented highway network with shared parameters among layers
(HWN-noRel). This is essentially a special case of $\model$ without
relational connections. 

\subsection{Experiment Settings}

We report three variants of the $\model$: a basic version that uses
standard Feedforward Neural Network as mini-column ($\model$-FNN)
and two versions of $\model$-HWN that use highway nets with shared
parameters ($\model$-HWN-full for full-batch mode and $\model$-HWN-mini
for mini-batch mode, as described in Sections~\ref{subsec:Highway-Network-as-Column},
\ref{subsec:Parameter-Sharing-for-Compactness} and \ref{subsec:Training-with-mini-batch}).
All neural nets use ReLU in the hidden layers.

Dropout is applied before and after the recurrent layers of $\model$-HWNs
and at every hidden layers of $\model$-FNN. Each dataset is divided
into 3 separated sets: training, validation and test sets. For hyper-parameter
tuning, we search for (i) number of hidden layers: 2, 6, 10, ...,
30, (ii) hidden dimensions, and (iii) optimizers: Adam or RMSprop.
$\model$-FNN has 2 hidden layers and the same hidden dimension with
$\model$-HWN so that the two models have equal number of parameters.
The best training setting is chosen by the validation set and the
results of the test set are reported. The result of each setting is
reported by the mean result of 5 runs. Code for our model can be found
on Github \footnote{https://github.com/trangptm/Column\_networks}

\subsection{Software Delay Prediction}

This task is to predict potential delay for an \emph{issue}, which
is an unit of task in an iterative software development lifecycle
\cite{choetkiertikul2015predicting}. The prediction point is when
issue planning has completed. Due to the dependencies between issues,
the prediction of delay for an issue must take into account all related
issues. We use the largest dataset reported in \cite{choetkiertikul2015predicting},
the JBoss, which contains 8,206 issues. Each issue is a vector of
15 features and connects to other issues through 12 relations (unidirectional
such as \texttt{blocked-by} or bidirectional such as \texttt{same-developer}).
The task is to predict whether a software issue is at risk of getting
delays (i.e., binary classification). 

\begin{figure}[h]
\centering{}\includegraphics[bb=0bp 105bp 675bp 440bp,clip,width=0.95\columnwidth]{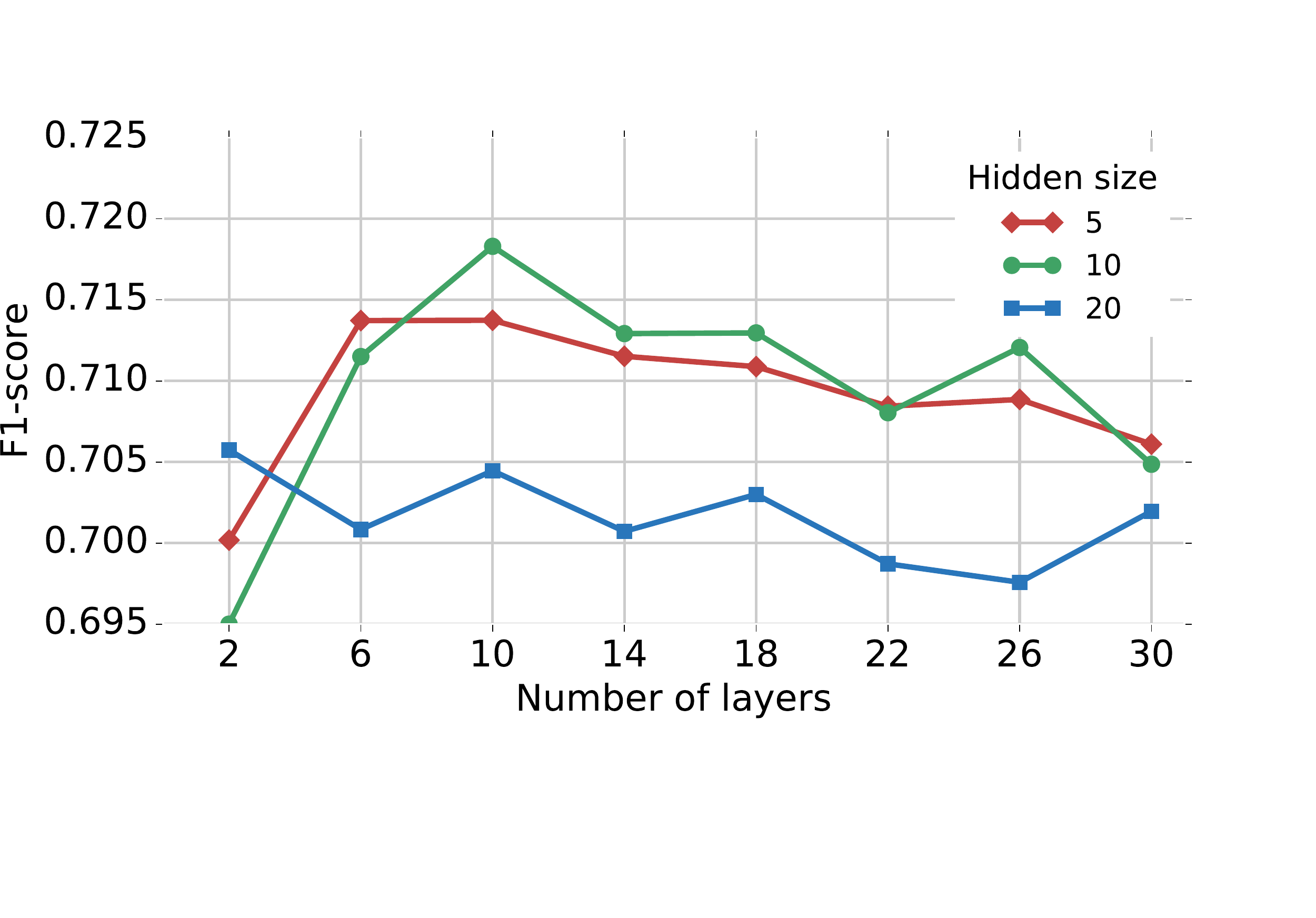}\caption{Performance (F1-score) of $\protect\model$-HWN on Software delay
prediction task with different numbers of layers and hidden sizes
\label{fig:Results-RHWN-layers}}
\end{figure}

Fig.~\ref{fig:Results-RHWN-layers} visualizes $\model$-HWN-full
performance with different numbers of layers ranging from 2 to 30
and hidden dimensions from 5, 10 to 20. The F1-score peaks at 10 hidden
layers and dimension size of 10.

Table~\ref{tab:Results-on-software} reports the F1-scores of all
methods. The two best classifiers in NetKit are wvRN-IC and wvRN-RL.
The non-collective HWN-noRel works surprisingly well \textendash{}
almost reaching the performance of the best collective SL-RF with
2 points short. This demonstrates that deep neural nets are highly
competitive in this domain, and to the best of our knowledge, this
fact has not been established. $\model$-HWN-full beats the best collective-method,
the SL-RF by 3.1 points. We lost 0.7\% in mini-batch training mode
but the gain of training speed was substantial - roughly 6x.

\begin{table}[h]
\centering{}%
\begin{tabular}{|lc||lc|}
\hline 
\emph{Non-neural} & \emph{F1} & \emph{Neural net} & \emph{F1}\tabularnewline
\hline 
\hline 
wvRN-IC & 54.7 & HWN-noRel & 66.8\tabularnewline
wvRN-RL & 55.8 & $\model$-FNN & 70.5\tabularnewline
\hline 
SL-LR & 65.3 & \textbf{$\model$-}HWN-full & \textbf{71.9}\tabularnewline
SL-RF({*}) & 68.8 & $\model$-HWN-mini & 71.2\tabularnewline
\hline 
\end{tabular}\caption{Software delay prediction performance. ({*}) Result reported in \cite{choetkiertikul2015predicting}.
\label{tab:Results-on-software}}
\end{table}

\subsection{PubMed Publication Classification}

We used the Pubmed Diabetes dataset consisting of 19,717 scientific
publications and 44,338 citation links among them\footnote{Download: http://linqs.umiacs.umd.edu/projects//projects/lbc/}.
Each publication is described by a TF/IDF weighted word vector from
a dictionary which consists of 500 unique words. We conducted experiments
of classifying each publication into one of three classes: Diabetes
Melitus - Experimental, Diabetes Melitus type 1, and Diabetes Mellitus
type 2. 

\subsubsection*{Visualization of hidden layers}

\begin{figure}[h]
\centering{}\includegraphics[bb=40bp 35bp 670bp 390bp,clip,width=0.97\columnwidth]{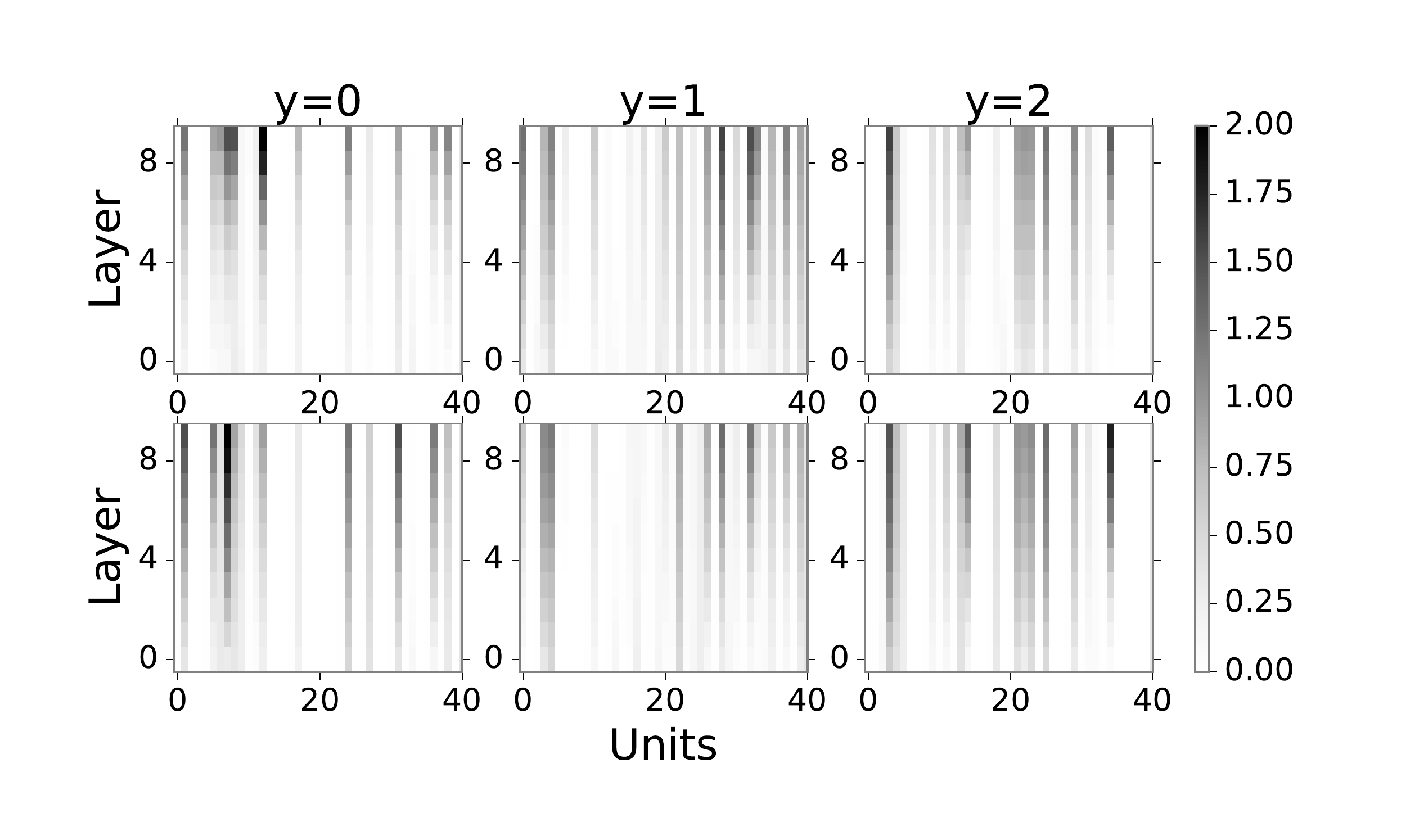}\caption{The dynamics of activations of 40 ReLU units through 10 hidden layers
of 2x3 samples - each column is for a class. \label{fig:Results-hidden-pubmed}}
\end{figure}

We randomly picked 2 samples of each class and visualized their ReLU
units activations through 10 layers of the $\model$-HWN (Fig.~\ref{fig:Results-hidden-pubmed}).
Interestingly, the activation strength seems to grow with higher layers,
suggesting that learn features are more discriminative as they are
getting closer to the outcomes. For each class a number of hidden
units is turned off in every layer. Figures of samples in the same
class have similar patterns while figures of samples from different
classes are very different. 

\subsubsection*{Classification accuracy}

The best setting for \textbf{$\model$}-HWN is with 40 hidden dimensions
and 10 recurrent layers. Results are measured in MicroF1-score and
MacroF1-score (See Table~\ref{tab:Results-on-diabetes}). The non-relational
highway net (HWN-noRel) outperforms two best baselines from NetKit.
The two version of \textbf{$\model$}-HWN perform best in both F1-score
measures. 

\begin{table}[h]
\centering{}%
\begin{tabular}{|lcc|}
\hline 
\emph{Method} & \emph{MicroF1-score} & \emph{MacroF1-score}\tabularnewline
\hline 
\hline 
wvRN-IC & 82.6 & 81.4\tabularnewline
wvRN-RL & 82.4 & 81.2\tabularnewline
\hline 
SL-LR & 88.2 & 87.9\tabularnewline
\hline 
HWN-noRel & 87.9 & 87.9\tabularnewline
$\model$-FNN & 89.4 & 89.2\tabularnewline
$\model$-HWN-full & \textbf{89.8} & \textbf{89.6}\tabularnewline
\textbf{$\model$-}HWN-mini & \textbf{89.8} & \textbf{89.6}\tabularnewline
\hline 
\end{tabular}\caption{Pubmed Diabetes classification results measured by MicroF1-score and
MacroF1-score. \label{tab:Results-on-diabetes}}
\end{table}

\subsection{Film Genre Prediction}

We used the MovieLens Latest Dataset \cite{harper2016movielens} which
consists of 33,000 movies. The task is to predict genres for each
movie given plot summary. Local features were extracted from movie
plot summary downloaded from IMDB database\footnote{http://www.imdb.com}.
After removing all movies without plot summary, the dataset remains
18,352 movies. Each movie is described by a Bag-of-Words vector of
1,000 most frequent words. Relations between movies are (\texttt{same-actor,
same-director}). To create a rather balanced dataset, 20 genres are
collapsed into 9 labels: (1) Drama, (2) Comedy, (3) Horror + Thriller,
(4) Adventure + Action, (5) Mystery + Crime + Film-Noir, (6) Romance,
(7) Western + War + Documentary, (8) Musical + Animation + Children,
and (9) Fantasy + Sci-Fi. The frequencies of 9 labels are reported
in Table~\ref{tab:Labels-freq}.

\begin{table}[h]
\centering{}%
\begin{tabular}{|lcccc|c|}
\hline 
\emph{Label} & \emph{0} & \emph{1} & \emph{2} & \multicolumn{1}{c}{\emph{3}} & \emph{4}\tabularnewline
\hline 
Freq(\%) & \emph{46.3} & \emph{32.5} & \emph{24.0} & \multicolumn{1}{c}{\emph{19.1}} & \emph{15.9}\tabularnewline
\hline 
\hline 
\emph{Label} & \emph{5} & \emph{6} & \emph{7} & \emph{8} & \multicolumn{1}{c}{}\tabularnewline
\cline{1-5} 
Freq(\%) & \emph{16.5} & \emph{14.8} & \emph{10.4} & \emph{11.5} & \multicolumn{1}{c}{}\tabularnewline
\cline{1-5} 
\end{tabular}\caption{The frequencies of 9 collapsed labels on Movielens \label{tab:Labels-freq}}
\end{table}

On this dataset, $\model$-HWNs work best with 30 hidden dimensions
and 10 recurrent layers. Table~\ref{tab:Results-on-genre-prediction}
reports the F-scores. The two best settings with NetKit are nbC-IC
and nbC-RL. $\model$-FNN performs well on Micro-F1 but fails to improve
MacroF1-score of prediction. \textbf{$\model$-}HWN-mini outperforms
$\model$-HWN-full by 1.3 points on Macro-F1.

\begin{table}[h]
\centering{}%
\begin{tabular}{|lcc|}
\hline 
\emph{Method} & \emph{Micro-F1} & \emph{Macro-F1}\tabularnewline
\hline 
\hline 
nbC-IC & 46.6 & 38.0\tabularnewline
nbC-RL & 43.5 & 40.4\tabularnewline
\hline 
SL-LR & 53.4 & 48.9\tabularnewline
\hline 
HWN-noRel & 50.8 & 45.2\tabularnewline
$\model$-FNN & 54.3 & 41.8\tabularnewline
\textbf{$\model$}-HWN-full & 57.4 & 52.7\tabularnewline
$\model$-HWN-mini & \textbf{57.5} & \textbf{54.1}\tabularnewline
\hline 
\end{tabular}\caption{Movie Genre Classification Performance reported in MicroF1 score and
MacroF1 score. \label{tab:Results-on-genre-prediction}}
\end{table}

\begin{figure}[h]
\centering{}\includegraphics[bb=20bp 100bp 660bp 440bp,clip,width=0.99\columnwidth]{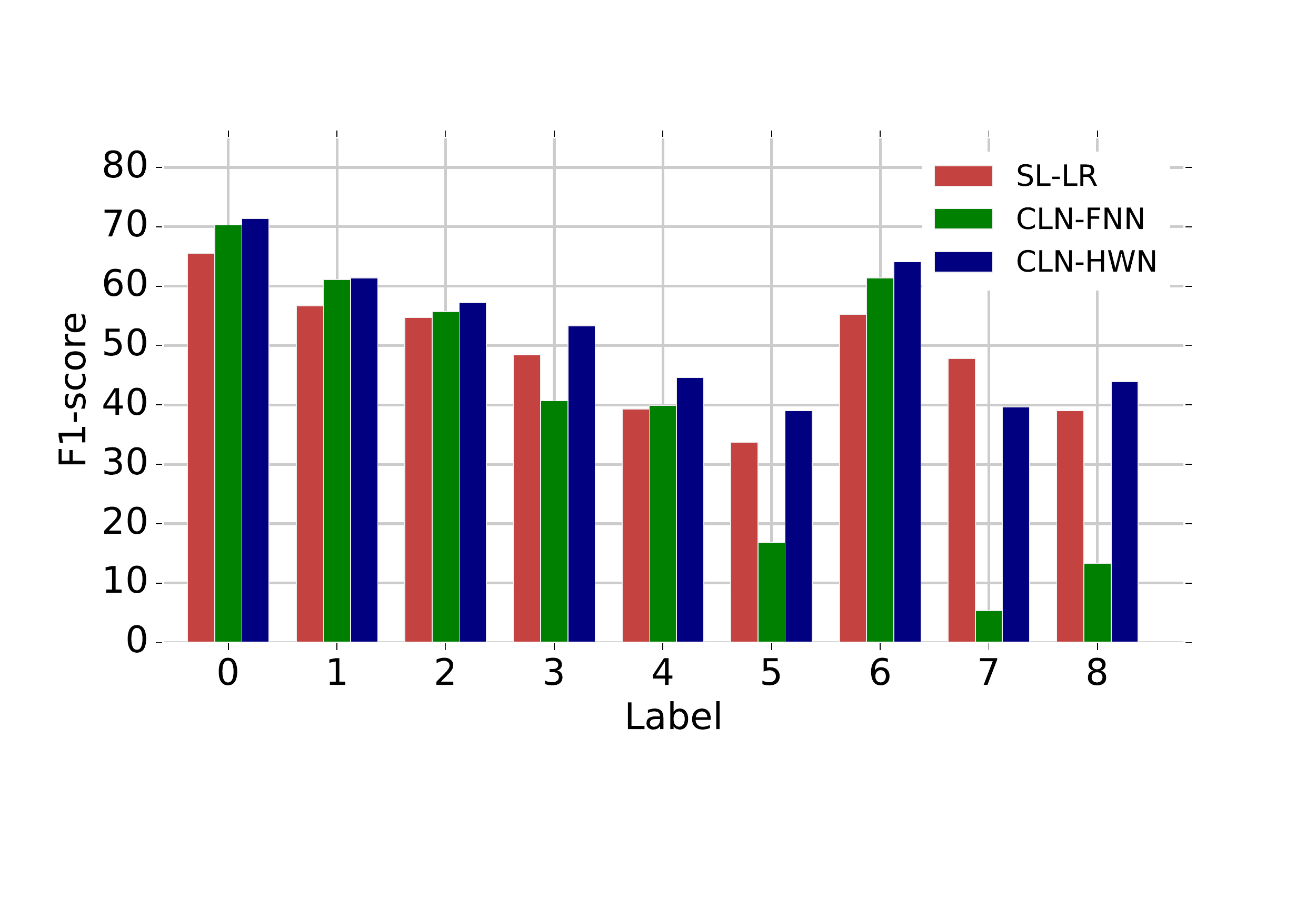}\caption{Genre prediction F1-score of SL-LR,$\protect\model$-FNN, $\protect\model$-HWN
on each label. Best viewed in color. \label{fig:Result-f1-each-label}}
\end{figure}

Fig.~\ref{fig:Result-f1-each-label} shows why $\model$-FNN performs
badly on MacroF1 (MacroF1 is the average of all classes' F1-scores).
While $\model$-FNN works well with balanced classes (in the first
three classes, its performance is nearly as good as \textbf{$\model$}-HWN),
it fails to handle imbalanced classes (See Table~\ref{tab:Labels-freq}
for label frequencies). For example, F1-score is only 5.4\% for label
7 and 13.3\% for label 8. In contrast, \textbf{$\model$}-HWN performs
well on all classes.

\section{Related Work \label{sec:Related-Work} }

This paper sits at the intersection of two recent independently developed
areas: Statistical Relational Learning (SRL) and Deep Learning (DL).
Started in the late 1990s, SRL has advanced significantly with noticeable
works such as Probabilistic Relational Models \cite{getoor1999upr},
Conditional Random Fields \cite{lafferty01conditional}, Relational
Markov Network \cite{Taskar-et-alUAI02} and Markov Logic Networks
\cite{Richardson-Domingos-ML06}. Collective classification is a canonical
task in SRL, also known in various forms as structured prediction
\cite{dietterich2008structured} and classification on networked data
\cite{macskassy2007cnd}. 

Two key components of collective classifiers are relational classifier
and collective inference \cite{macskassy2007cnd}. \emph{Relational
classifier} makes use of predicted classes (or class probabilities)
of entities from neighbors as features. Examples are wvRN \cite{macskassy2007cnd},
logistic based \cite{domke2013structured} or stacked graphical learning
\cite{choetkiertikul2015predicting,kou2007stacked}. \emph{Collective
inference} is the task of jointly inferring labels for entities. This
is a subject of AI with abundance of solutions including message passing
algorithms \cite{Pearl88}, variational mean-field \cite{opper2001advanced}
and discrete optimization \cite{truyen2014tree}. Among existing
collective classifiers, the closest to ours is stacked graphical learning
where collective inference is bypassed through stacking \cite{kou2007stacked,yu2010sequential}.
The idea is based on learning a stack of models that take intermediate
prediction of neighborhood into account. 

The other area is Deep Learning (DL), where the current wave has offered
compact and efficient ways to build multilayered networks for function
approximation (via feedforward networks) and program construction
(via recurrent networks) \cite{lecun2015deep,schmidhuber2015deep}.
However, much less attention has been paid to general networked data
\cite{monner2013recurrent}, although there has been work on pairing
structured outputs with deep networks \cite{belanger2015structured,do2010neural,tompson2014joint,yu2010sequential,zheng2015conditional}.
Parameter sharing in feedforward networks was recently analyzed in
\cite{liao2016bridging,pham2016faster}. The sharing eventually transforms
the networks in to recurrent neural networks (RNNs) with only one
input at the first layer. The empirical findings were that the performance
is good despite the compactness of the model. Among deep neural nets,
the closest to our work is RNCC model \cite{monner2013recurrent},
which also aims at collective classification using RNNs. There are
substantial differences, however. RNCC shuffles neighbors of an entities
to a random sequence and uses \emph{horizontal RNN} to integrate the
sequence of neighbors. Ours emphasizes on vertical depth, where parameter
sharing gives rise to the \emph{vertical RNNs}. Ours is conceptually
simpler \textendash{} all nodes are trained simultaneously, not separately
as in RNCC.

\section{Discussion \label{sec:Discussion}}

This paper has proposed Column Network ($\model$), a deep neural
network with an emphasis on fast and accurate collective classification.
$\model$ has linear complexity in data size and number of relations
in both training and inference. Empirically, $\model$ demonstrates
a competitive performance against rival collective classifiers on
three real-world applications: (a) delay prediction in software projects,
(b) PubMed Diabetes publication classification and (c) film genre
classification. 

As the name suggests, $\model$ is a network of narrow deep networks,
where each layer is extended to incorporate as input the preceding
neighbor layers. It somewhat resembles the columnar structure in neocortex
\cite{mountcastle1997columnar}, where each narrow deep network plays
a role of a mini-column. We wish to emphasize that although we use
highway networks in actual implementation due to its excellent performance
\cite{pham2016faster,srivastava2015training,tran2016choiceb}, \emph{any
feedforward networks can be potentially be used} in our architecture.
When \emph{parameter sharing} is used, the feedforward networks become
recurrent networks, and $\model$ becomes a network of interacting
RNNs. Indeed, the entire network can be collapsed into a giant feedforward
network with $n-n$ input/output mappings. When relations are shared
among all nodes across the network, $\model$ enables \emph{translation
invariance} across the network, similar to those in CNN. However,
the $\model$ is not limited to a single network with shared relations.
Alternatively, networks can be IID according to some distribution
and this allows relations to be specific to nodes.

There are open rooms for future work. One extension is to learn the
pooling operation using attention mechanisms. We have considered
only homogeneous prediction tasks here, assuming instances are of
the same type. However, the same framework can be easily extended
to multiple instance types.

\section*{Acknowledgement}

Dinh Phung is partially supported by the Australian Research Council
under the Discovery Project DP150100031

\bibliographystyle{aaai}
\bibliography{ME,trang,truyen}

\end{document}